\documentclass{ieeeaccess}
\usepackage{cite}
\usepackage{amsmath,amssymb,amsfonts}
\usepackage[numbers]{natbib}
\usepackage{algorithmic}
\usepackage{graphicx}
\usepackage{float}
\usepackage{hyperref}
\usepackage{textcomp}
\usepackage{tabularx}
\usepackage{array}
\usepackage{multirow}
\newcolumntype{P}[1]{>{\centering\arraybackslash}p{#1}}

\newcommand{\headline[1]}{\vspace*{5pt}
	\noindent
	\textbf{\textit{#1}}
	\vspace{5pt}}

\def\BibTeX{{\rm B\kern-.05em{\sc i\kern-.025em b}\kern-.08em
    T\kern-.1667em\lower.7ex\hbox{E}\kern-.125emX}}
\begin{document}

\history{Date of publication xxxx 00, 0000, date of current version xxxx 00, 0000.}
\doi{00.0000/ACCESS.2024.DOI}

\title{Multi-step feature fusion for natural disaster damage assessment on satellite images}
\author{\uppercase{Mateusz \.Zarski}\authorrefmark{1} and
\uppercase{Jaros\L aw A. Miszczak\authorrefmark{1}
\address[1]{Institute of Theoretical and Applied Informatics, Polish Academy of Sciences, Bałtycka 5, Gliwice, 44-100, Poland}}}

\markboth
{Żarski \headeretal: Multi-step feature fusion...}
{Żarski \headeretal: Multi-step feature fusion...}

\corresp{Corresponding author: Mateusz \.Zarski (e-mail: mzarski@iitis.pl).}

\begin{abstract}
Quick and accurate assessment of the damage state of buildings after natural disasters is crucial for undertaking properly targeted rescue and subsequent recovery operations, which can have a major impact on the safety of victims and the cost of disaster recovery. The quality of such a process can be significantly improved by harnessing the potential of machine learning methods in computer vision. This paper presents a novel damage assessment method using an original multi-step feature fusion network for the classification of the damage state of buildings based on pre- and post-disaster large-scale satellite images. We introduce a novel convolutional neural network (CNN) module that performs feature fusion at multiple network levels between pre- and post-disaster images in the horizontal and vertical directions of CNN network. An additional network element -- \textit{Fuse Module} -- was proposed to adapt any CNN model to analyze image pairs in the issue of pair classification. We use, open, large-scale datasets (IDA-BD and xView2) to verify, that the proposed method is suitable to improve on existing state-of-the-art architectures. We report over a 3 percentage point increase in the accuracy of the Vision Transformer model.
\end{abstract}


\titlepgskip=-25pt

\maketitle

\section{Introduction}
\label{sec:introduction}
\PARstart{N}{atural} disasters, such as earthquakes, widespread fires, hurricanes, floods, etc., are characterized by the suddenness of their occurrence, unpredictability, great danger to the residents of the affected area, severe damage to infrastructure, and ability to cover vast areas of urbanized terrain. Over the past 30 years, year-on-year, there has been a noticeable increase in their reported occurrence and a nearly twofold increase in the economic losses incurred as a result of their occurrence~\cite{natural_disaster_book}, however many pre-2000 natural disasters and their outcomes remain undocumented~\cite{owid-disaster-database-limitations}. Yearly, natural disasters take a toll of over 60 thousand lives and affect the lives of over 150 million people. Considering the data from year 2023, the most vulnerable areas are Asia (about 40\% of all natural disasters) and South America (25\% of all disasters). Although floods and violent storms account for the vast majority of natural disasters that occur (more than 75\% of all disasters if combined), earthquakes, which account for only 8\% of all natural disasters, claim the most lives (approximately 74\% of all fatalities), with a single occurrence (February 2023 Turkey–Syria earthquake) costing the lives of as many as 60,000 people. They are also the second-largest cause of property damage associated with natural disasters, after violent storms~\cite{2023_disasters}.

For this reason, current related scientific work emphasizes specific elements related to disaster impact mitigation, such as simulation and prediction of disaster impacts, the development of emergency management and planning methods in the event of disaster occurrence, also in scope of past disasters, and methods for assessing their impact.

Accurately predicting natural disasters or their triggers based on environmental indicators remains a challenging task. Researchers proposed the use of structural prediction methods~\cite{struct_pred} or Group Method of Data Handling~\cite{Moosavi2019} to predict regions of increased precipitation resulting in flooding with various levels of accuracy. Other scientific work proposes the development of systems for collecting and assessing multimodal data for identifying and assessing disaster-related risks~\cite{disaster_monitoring}. Disaster impact prediction works also describe the potential impact of disasters on population and infrastructure~\cite{R_Ginantra_2021}, consider economic indicators~\cite{su11030868}, and predict mobility problems~\cite{mobility} in the event of a disaster.

In the research related to emergency management and response planning, one of the methods employed in such activities by researchers focuses on general solutions for emergency response planning using lecture and tabletop simulation techniques ~\cite{gunawan2019improving, mahdi_sim}. Other focus on the optimal allocation of resources and the management of response measures~\cite{WEX2014697, CHAKRAVARTY20113}. Additionally, some of the studies focus on specific natural disasters, such as Hurricane Katrina~\cite{Perry,McSwain}.

The last of the scientific work domains related to natural disasters is the development of methods for its impact assessment. These methods use different data modalities in research work. The latest trend is to use the integration of information in the form of images and text from multiple sources, with social media as the main source of information~\cite{Chen2021, Kryvasheyeu2016, Nguyen2017}. Other researchers focus on the use of UAV or other aerial image data~\cite{9553712, 9377916, Zhu_2021_WACV}. Some research effort was also made to gather reliable image data for benchmarks in UAV applications~\cite{Rahnemoonfar_2023} or use synthetic data for algorithms training~\cite{9913977}. There are also a number of works that use satellite images in various modalities for this purpose -- \textit{e.g.}~\cite{YAMAZAKI2007, Gillespie, nhess-8-707-2008}.

In this article, we focus on expanding the methods of natural disaster impact assessment by addressing the problem of efficient analysis of vision data from pairs of satellite images from before and immediately after a natural disaster. The rest of the article is organized as follows: in Section~\ref{sec:related} we describe other research that is connected to our method. Section~\ref{sec:method} describes our \textit{Fuse Module} and the details of CNN architecture we developed specifically for the task, as well as the details of its implementation. In Section~\ref{sec:exp_setup} we describe our experimental setup with datasets used, and network architectures tested, and in Section~\ref{sec:evaluation} we describe the tests we performed and assess the results. Finally, in Section~\ref{sec:conclusions} we provide our final remarks and discuss the possible future research directions.

\section{Related works}\label{sec:related}

In our work, we use satellite vision data in the feature fusion paradigm to develop a deep learning model capable of assessing the damage state of infrastructure with image pairs -- from before and after the occurrence of a natural disaster. 

As stated in Section~\ref{sec:introduction}, there are a number of methods focused on using satellite visual data for this task. Research using deep learning in computer vision employed relatively shallow CNNs to assess the condition of buildings~\cite{7730352, Dotel2020DisasterAF} and also two-step classifiers for semantic segmentation and further damage assessment of objects in single images~\cite{alisjahbana2024deepdamagenet}. Newer implementations of deep learning methods also use transformer networks for a similar task~\cite{Kaur2023}. Other researchers focus on utilizing other spectra in satellite image classification~\cite{inbook_ms}. Authors also note the problem of imbalance in satellite training data~\cite{WANG2022104328}, which we also address in our research. To focus on a single issue related to evaluating facilities after natural disasters, we decided to use large-scale RGB data only. In our research, we also chose to skip the repeatedly solved task of segmenting buildings from satellite images (\cite{jiwani2021semantic, rs11040403, rs12101544, 9412295} \textit{etc.}), since this step of analysis is not relevant to our solution.

The problem of assessing consecutive changes in the state of the building is closely related to time series analysis with visual data~\cite{10.1117/12.2309486}, with such analyses also performed on satellite images~\cite{GUYET201617}. One of the notable approaches is the use of multi- and hyperspecral data in the interpretation and segmentation of remote sensing images (RSIs). In this area of research, authors used the exchange of CNN and transformer network features~\cite{LiXin_cnn_trans} and attention maps fusion~\cite{rs15235610}. Other works on the attention mechanism for RSI data also include a synergistic attention module allowing to preserve spatial details in channel-wise attention mechanisms~\cite{10041998}. In structural engineering, such an approach was used in monitoring newly constructed building areas~\cite{HUANG2020111802} and the detection of damaged building regions~\cite{ijgi6050131}. Other modalities of data were also used, such as SAR (synthetic aperture radar) satellite images, for similar purposes~\cite{SAR1, 7845580}. However, due to the data scarcity, most of the time series-related research in building damage state assessment focuses on using forced vibrations~\cite{10.1115/1.2718241, vibration2}. 

For this purpose, a more preferable approach is using image pairs in such analysis, which, on the other hand, is related to the image verification problem~\cite{appalaraju2018image} often used in \textit{e.g.} security access systems for face or fingerprint verification. In civil engineering, verification was used in two approaches. In the first one, researchers use siamese-type networks with CNN architectures such as U-Net, utilizing a single network branch for two images~\cite{9554054, rs13050905}. Other researchers focus on using feature or temporal fusion and multiple-branch networks, each for one of the images~\cite{9412295, weber2020building} and combining features at the classification stage.  

Feature fusion in CNNs is commonly performed in a single step after a certain stage of the network, and such approaches can mostly be divided into early and late-feature fusion. Early feature fusion networks, as in~\cite{Chen2023} perform the fusion after the initial CNN part of the transformer model, while late feature fusion networks~\cite{FF9200676, TANG2017188} fuse the features after a certain stage in the network, usually before the classifier. Furthermore, for the fusion itself, most of the time simple concatenation of features is employed~\cite{DING2022246, FF28237195} which considerably increases the feature space and thus increases the network's parameter count.

In our research, we target the feature fusion approach, expanding on it by introducing multi-step feature fusion with our \textit{Fuse Module} that introduces microarchitectures to sequential or pre-defined CNN models. We also inspect various operations to perform the fusion that allows the network to maintain a low parameter count for faster inference. We test our approach on common CNN architectures as well as introducing it to transformer-type networks.

\section{Method}\label{sec:method}

This work proposes a novel method of performing feature fusion at multiple stages of the network. It incorporates the \textit{Fuse Module} into the network architecture (see Fig.~\ref{fig:architecture}) trained for the task of assessing the building damage state on satellite image pairs after natural disasters. The \textit{Fuse Module} allows for the exchange and comparison of information based on different features extracted from a pair of images at different stages of processing through the network. It can fuse information both horizontally and vertically through the network by performing simple operations described later in the paper. To distinguish the direction relative to the network on which the Fuse Module operates, we have distinguished between its two types. \textit{Fuse Module H} -- which operates horizontally, performing feature fusion between the same stages of the network, and \textit{Fuse Module V}, fusing features between two successive stages of the network. Both modules can perform different operations, but the principle of their function remains the same (see Fig.~\ref{fig:fuse_module} for overall visual description). We expand on it later in Subsection~\ref{sec:implementation}.

\Figure[t!](topskip=0pt, botskip=0pt, midskip=0pt)[width=0.80\textwidth]{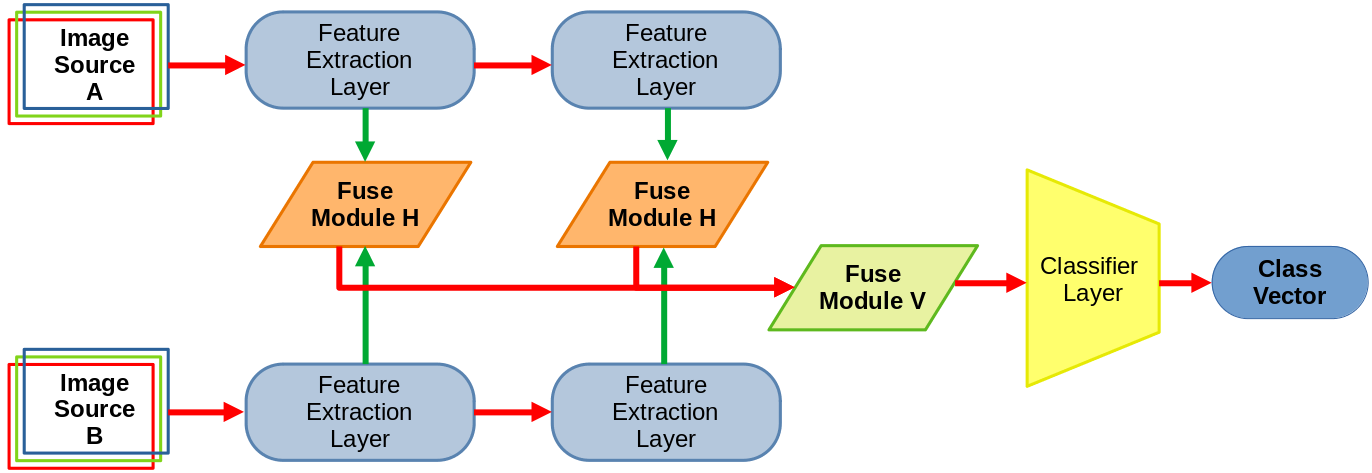}
{Example of network architecture using horizontal (\textit{Fuse H}) and vertical (\textit{Fuse V}) fusion. The flow of information through the network in the horizontal direction is marked with green lines, while the vertical direction is marked with red lines. Image Sources A and B are respectively pre- and post-natural disaster images registered on the same structure.\label{fig:architecture}}

\Figure[t!](topskip=0pt, botskip=0pt, midskip=0pt)[width=0.95\linewidth]{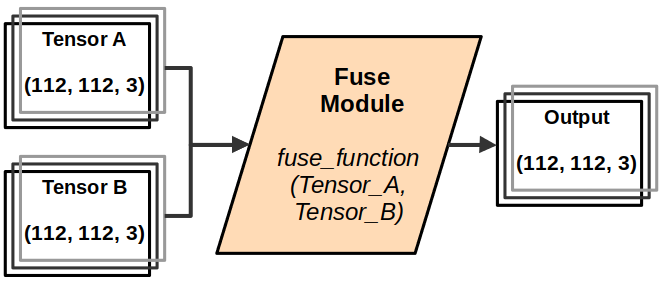}
{The key feature of the introduced \textit{Fuse Module} -- the ability to fuse information from multiple sources while maintaining initial feature space dimensions\label{fig:fuse_module}}

\subsection{Implementation}\label{sec:implementation}

\headline[Fusion functions]

In our work, we inspected various operations to perform the feature fusion, both efficiently in terms of the number of operations performed by the module and effectively in terms of maintaining the ability to train the model. In Tab.~\ref{tab:fuses} we provide all the fusion functions we tested in our work (for \textit{feature tensor} $A$ and $B$) and from now on (\textit{e.g.} in result tables in Section~\ref{sec:evaluation}), the fusion functions will be referenced with their numbers listed in the table. In our test setup in Section~\ref{sec:evaluation}, we cross-analysed pairs of the shown functions for their use in \textit{Fuse Module H} and \textit{Fuse Module V} in a single model at the same time, as well as their use in predefined models in horizontal-fusion only scenarios.

As depicted in Fig.~\ref{fig:architecture}, the fusion operation takes place after each feature extraction stage in the CNN model and before final classification layers. After each stage, the \textit{Fuse Module H} obtains features from two branches of the network and performs \textit{fuse\_function} on them (see Fig.~\ref{fig:fuse_module}). This allows for the exchange of the information between the branches of the CNN. Next, when next features are obtained from the CNN branches, after again fusing them with \textit{Fuse Module H}, the obtained fused tensors from two subsequent feature extractor stages are fused together with \textit{Fuse Module V} before final classification with MLP stage.

The implementation itself was performed with the PyTorch framework and was introduced in the training procedure of the network. We tested our implementation of the \textit{Fuse Module} for the time needed to perform the fusion operation and compared it to the network with a corresponding number of parameters, without the use of the \textit{Fuse Module}. The results of the comparison are depicted in Fig.~\ref{fig:inference_time}. With a sample size of $1000$ tensors, we found no significant increase in the inference time of the network with the \textit{Fuse Module} implemented, as due to the compact architecture of the network, the longest inference time with \textit{fuse function 3} still didn't exceed $0.42$~\textit{ms}. Note that the initial longer inference times for \textit{fuse function 1} and \textit{2} may be due to CUDA warm up period, and otherwise network without the \textit{Fuse Module} addition constantly performs the inference at $0.23$~\textit{ms} per image. Furthermore, inference time increase can still be lowered by employing only computations within GPU memory, \textit{e.g.} with JAX library.

\Figure[t!](topskip=0pt, botskip=0pt, midskip=0pt)[width=0.99\linewidth]{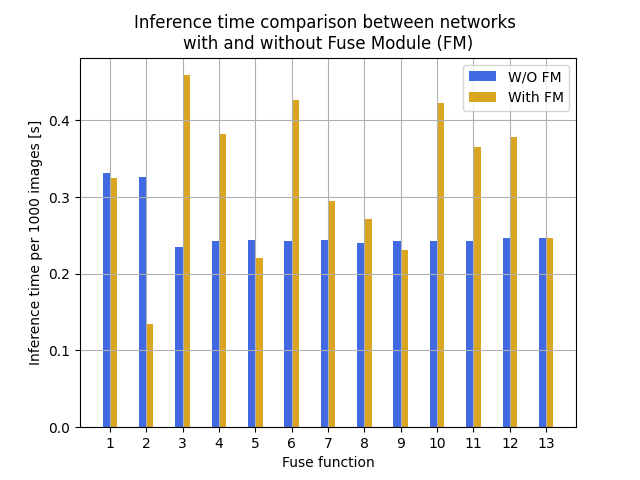}
{The comparison of inference times on 1000 images of network with and without the \textit{Fuse Module}\label{fig:inference_time}}

\begin{table}[]
\centering
	\caption{List of fuse functions implemented in \textit{Fuse Module} tested in our work}
	\vspace*{2pt}
\begin{tabular}{P{2cm}|P{4.8cm}}
Fuse Function & \begin{tabular}[c]{@{}c@{}}Operation performed on \\ $ft_{A}$ and $ft_{B}$ \end{tabular} \\[2ex] \hline \hline
1  & $ft_{A} \cdot ft_{B}$ \\[1ex]
2  & $\left | ft_{A} - ft_{B} \right |$ \\[1ex]
3  & $\sqrt{ft_{A}^2 - ft_{B}^2}$ \\[1ex]
4  & $ft_{A} @ ft_{B}$$^{*}$ \\[1ex]
5  & $ft_{A} + ft_{B}$ \\[1ex]
6  & $MEAN(ft_{A} \frown ft_{B})$$^{**}$ \\[1ex]
7  & $MAX(ft_{A} \frown ft_{B})$ \\[1ex]
8  & $MIN(ft_{A} \frown ft_{B})$ \\[1ex]
9  & $\left |ft_{A}^2 \cdot ft_{B}^2 \right |$ \\[1ex]
10 & $ft_{A} \bigotimes ft_{B}$ \\[1ex]
11 & $\sigma(ft_{A} \frown ft_{B})$ \\[1ex]
12 & $NORM(ft_{A} \frown ft_{B})$$^{***}$ \\[1ex]
13 & $\sigma^{2}(ft_{A} \frown ft_{B})$ \\[1ex]
\end{tabular}\label{tab:fuses}
\newline \raggedright $^{*}$ $@$ represents matrix batch multiplication
\newline \raggedright $^{**}$ $\frown$ represents matrix concatenation on extended, singular dimension
\newline \raggedright $^{***}$ matrix Frobenius norm
\end{table}

\headline[Loss Function]

To train the networks used in our work, we used the Mean Squared Error (MSE). However, due to the problem of a large unbalanced dataset highlighted earlier, we also added an element to the loss function to account for the large difference in class size. Ultimately, our loss function was
\begin{equation}
\mathcal{L}_{MSEw}(Y, \hat{Y}) = \frac{1}{N}\sum_{i=1}^{N}(Y_{i} - \hat{Y_{i}})\cdot W_{i},
\end{equation}
where $W_{i}$ represents the weight of each class for which the loss function is calculated, $N$ is the total count of examples in a single batch, and $Y_{i}$ and $\hat{Y_{i}}$ are successively the ground truth and the output from CNN.

Class weights were calculated separately for each training set from the datasets used in the paper. They were calculated using the following formula:
\begin{equation}
W_{i} = \frac{N_{tot}}{C \cdot N_{i}},
\end{equation}
where $N_{tot}$, is the total count of datapoints in the dataset, $C$ stands for the number of classes in the dataset, and $N_{i}$ is the number of class $i$ datapoints in the dataset.

In Section~\ref{sec:exp_setup} we provide additional information on the class imbalance and calculated class weights for the datasets used in our research.

\section{Experimental setup}\label{sec:exp_setup}

Two open source datasets were used in the article. \textbf{IDA-BD}~\cite{idabd} and \textbf{xView2}~\cite{xview}, both of which contain satellite images from before and after the natural disaster. IDA-BD contains 87 pre- and post-disaster images from the WorldView-2 satellite taken before and after Hurricane Ida in 2021 in Louisiana. xView2 dataset contains over 44 thousand images from a total of 17 natural disaster occurrences including earthquakes, tsunamis, floods, volcanic eruptions, winds, and wildfires. In addition to the images, both datasets also contain the semantic segmentation labels for buildings used in our work, which allow us to cut buildings out of the images, obtaining crops for CNN training. In both datasets, the degree of structural damage was rated on a four-point scale from 0 to 3, where 0 meant no damage and 3 meant complete destruction of the building. An example of images and cropped data points for training is shown in Fig.~\ref{fig:dataset}.

\Figure[t!](topskip=0pt, botskip=0pt, midskip=0pt)[width=0.65\textwidth]{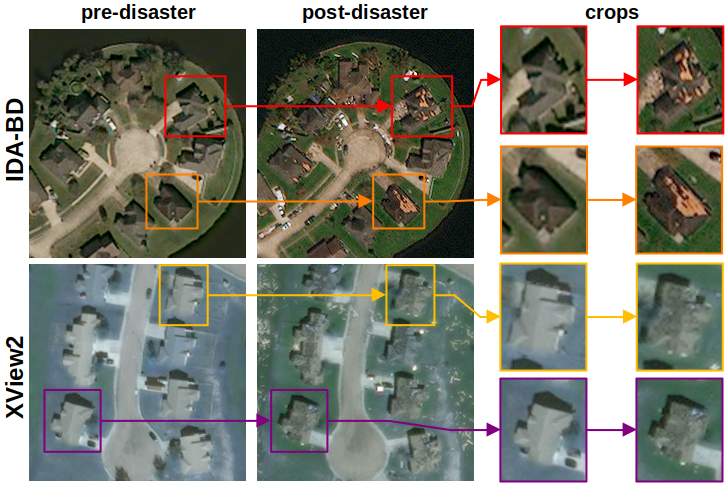}
{Example of images in IDA-BD and xView datasets from pre- and post-disaster along with crops selected for CNN training.\label{fig:dataset}}

To ensure that the data points used in the training included as much information as possible about the building rather than its surroundings, they were cropped with the information of the smallest area rectangular bounding box from the segmentation label images. The image was then transformed to a square using a warp perspective transformation and resized using quadratic interpolation to dimensions of 224 by 224 pixels. Additional augmentations were also implemented to the training images, including random flips, rotations, affine, Gaussian blur and Gaussian noise. Images from before the natural disaster constituted the \textit{Tensor A}, while images from after the disaster -- the \textit{Tensor B} during training of the algorithm. As for further preprocessing -- before the training images were normalized.

As stated in Section~\ref{sec:method}, the datasets were highly imbalanced in the way that most of the buildings were either undamaged or only with minor damages. For this reason, the weights corresponding to each damage class were counted separately for each of the training sets from the datasets, which were then used in the training.  A summary of the classes in the training sets and the weights for each class are shown in Tab.~\ref{tab:dataset}.

\begin{table}[]
\centering
	\caption{Class count breakdown and weights used during training for datasets used}
	\vspace*{2pt}
\begin{tabular}{cc|ccccc}
\multicolumn{2}{c|}{\textbf{Dataset}} &
  \textbf{\begin{tabular}[c]{@{}c@{}}Class \\ {[}0{]}\end{tabular}} &
  \textbf{\begin{tabular}[c]{@{}c@{}}Class \\ {[}1{]}\end{tabular}} &
  \textbf{\begin{tabular}[c]{@{}c@{}}Class \\ {[}2{]}\end{tabular}} &
  \textbf{\begin{tabular}[c]{@{}c@{}}Class \\ {[}3{]}\end{tabular}} &
  \textbf{Total} \rule[-2ex]{0pt}{5ex} \\ \hline \hline
                & Count    & 9749  & 3291  & 1148  & 42    & 14230  \rule[-2ex]{0pt}{5ex} \\[1ex]
 \textbf{IDA-BD} & {[}\%{]} & 68.51 & 23.13 & 8.07  & 0.30  & -      \\[1ex]
                & Weight   & 0.365 & 1.081 & 3.099 & 84.702 & -      \\[1ex] \hline
                & Count    & 97389 & 11754 & 12897 & 7940  & 129980 \rule[-2ex]{0pt}{5ex} \\[1ex]
 \textbf{xView2} & {[}\%{]} & 74.93 & 9.04  & 9.92  & 6.11  & -      \\[1ex]
                & Weight   & 0.334 & 2.765 & 2.520 & 4.093 & -     \\[1ex]
\end{tabular}\label{tab:dataset}
\end{table}

As can be seen, the two datasets differ significantly from each other. In the IDA-DB dataset, buildings described by class 3 (the most damaged) do not account for even 1\% of all buildings included in the dataset. In xView2, the most damaged buildings account for just over 6\%. For the other classes, except class 1 more heavily represented by the IDA-BD dataset, the differences are no longer so apparent.  

Differences in datasets also pose a problem in the issue of domain shift, i.e. when an algorithm trained on one dataset is applied to another, even when solving the same task. In addition, in the study, we had no control over the assignment of classes to buildings with given damage -- for this reason, it can be expected that buildings damaged similarly on different datasets would be described with a different damage class, as they were classified by separate teams. We plan on expanding on domain shift scenarios in our future work.

\section{Evaluation}\label{sec:evaluation}

In our evaluation experiments, we used several CNN architectures, including predefined networks for our baseline experiments, our own, simple, low-parameter CNN for fusion functions evaluation, and also predefined architectures modified by our fuse module to show the versatility of our method. In order to tune training hyperparameters, we performed grid search covering learning rate, augmentation methods and learning rate schedulers parameters. Our primary goal was to monitor accuracy and F1 score (as average F1 score for all classes to account for datasets imbalance) on testing datasets.

For all of the experiments, we used the Ubuntu 22.04 system with RTX 3080 GPU. Our programming environment was Python 3.10, and for all matrix operations, we used PyTorch 2.2. Our source code is available in a Github~\cite{eq_git} repository.

\subsection{Baseline experiments}\label{sec:baseline}

In our baseline experiments, we used predefined networks in a common setup for the feature fusion task -- we allowed the network to extract features from both pre- and post-disaster images and concatenated them before the classifier layers. The networks were trained for 60 epochs with 5 different seeds, although the number of epochs could be reduced to 45-50. We also used \textit{Reduce on Plateau} scheduler for learning rate management and ADAM optimizer. The results of our baseline experiments are shown as mean values from 5 seeds in Tab.~\ref{tab:baseline}

\begin{table}[]
\centering
\caption{Baseline experiments results}
\vspace*{2pt}
\begin{tabular}{c|cccc}
\textbf{Network} &
  \textbf{Dataset} &
  \textbf{\begin{tabular}[c]{@{}c@{}}Training\\ Accuracy\\ {[}\%{]}\end{tabular}} &
  \textbf{\begin{tabular}[c]{@{}c@{}}Testing\\ Accuracy\\ {[}\%{]} / \\ F1 score {[}\%{]}\end{tabular}} &
  \textbf{\begin{tabular}[c]{@{}c@{}}Parameter\\ Count {[}M{]}\end{tabular}} \\[1ex] \hline \hline 
\multirow{2}{*}{\textbf{\begin{tabular}[c]{@{}c@{}}VGG\\ 16\end{tabular}}}    & IDA-BD & 99.17 & 88.20 / 70.88          & \multirow{2}{*}{237.04}  \rule[-2ex]{0pt}{5ex}       \\
                                                                              & xView2 & 89.43 & \textbf{88.14} / \textbf{70.20} &                                 \\ \hline
\multirow{2}{*}{\textbf{\begin{tabular}[c]{@{}c@{}}ResNet\\ 18\end{tabular}}} & IDA-BD & 99.95 & 88.58 / 71.02         & \multirow{2}{*}{\textbf{11.18}} \rule[-2ex]{0pt}{5ex} \\
                                                                              & xView2 & 90.11 & 87.57 / 70.06         &                                 \\ \hline
\multirow{2}{*}{\textbf{\begin{tabular}[c]{@{}c@{}}ResNet\\ 50\end{tabular}}} & IDA-BD & 93.56 & \textbf{88.61} / \textbf{71.16} & \multirow{2}{*}{23.52}    \rule[-2ex]{0pt}{5ex}      \\
                                                                              & xView2 & 89.09 & 87.40 / 70.03         &                                 \\ \hline
\multirow{2}{*}{\textbf{\begin{tabular}[c]{@{}c@{}}ViT\\ B\end{tabular}}}     & IDA-BD & 91.70 & 88.51 / 70.00         & \multirow{2}{*}{85.80}     \rule[-2ex]{0pt}{5ex}     \\
                                                                              & xView2 & 89.90 & 86.31 / 69.93        &                                 \\ \hline
\multirow{2}{*}{\textbf{\begin{tabular}[c]{@{}c@{}}ViT\\ L\end{tabular}}}     & IDA-BD & 88.44 & 87.08 / 69.59         & \multirow{2}{*}{303.31}    \rule[-2ex]{0pt}{5ex}     \\
                                                                              & xView2 & 90.03 & 86.50 / 69.94        &                                
\end{tabular}\label{tab:baseline}
\end{table}

As can be seen from the results of our baseline tests, there is only a slight difference between the results of the algorithms on the IDA-BD and xView2 datasets, in addition, it is more noticeable when comparing the results on the training datasets rather than the test datasets. Also surprising are the relatively low results achieved by transformer-type networks -- ViT versions B and L. This may be due to the low dimensions or level of detail of the images in the dataset. In the baseline tests, again surprisingly, the best results were achieved by the VGG16 and ResNet50 networks, however considering the parameter count, and relative novelty of the architectures, only ResNet50, and the ViT B transformer were chosen for further testing of the \textit{Fuse Module}. Additional attention should also be focused on the F1 score metric -- using datasets with such a high imbalance, a visibly lower F1 score metric is expected, but this may be a reason to further revise the loss function in future work, as it was not the main scope of this paper.

\subsection{Fusion Functions Evaluation}\label{sec:fusion_eval}

In the first part of the evaluation of our proposed data tensor fusion functions, we conducted an analysis separately for both types of fusion direction using all 13 methods from Tab~\ref{tab:fuses} on the IDA-BD dataset. We first checked only fusion in the horizontal direction using \textit{Fuse Module H} in our \textit{Fuse\_H} architecture to connect network segments at the same depth. Then we performed a similar check using only vertical fusion with \textit{Fuse Module V} in \textit{Fuse\_V} architecture -- which in this case works analogously to the residual connection from the ResNet family of architectures. Finally, in the \textit{Fuse\_HV} architecture, we used a combination of both modules and verified the performance of all fusion functions by cross-analyzing their application in the same model. For the fairness of comparison with baseline experiments, for our subsequent experiments, we used the same training recipe, scheduler, and optimizer as in baseline. Below, we briefly describe the architectures of the networks used and what results we obtained.

\headline[Horizontal Fusion]

In horizontal fusion experiments, we used our \textit{Fuse\_H} architecture with 4 separate feature extraction stages for \textit{Tensor A} and \textit{Tensor B} performed with convolutional layers and the features were fused horizontally after each stage. We also added two more feature extraction layers after the final fusion by concatenation of the features, followed by a 4-layer MLP classifier. For nonlinearities within the architecture, ReLU function was used. 

The network had a total of \textbf{$28.94$} M parameters which is close to the parameter count of the ResNet50 model and much lower than VGG or ViT implementations. In Tab.~\ref{tab:horizontal} we show 5 best performing fusion methods for our \textit{Fuse\_H} model.

\begin{table}[]
\centering
	\caption{Five best performing methods in our \textit{Fuse\_H} model}
	\vspace*{2pt}
\begin{tabular}{P{1.2cm}|P{1.7cm}P{1.7cm}P{1.7cm}}
\textbf{\begin{tabular}[c]{@{}c@{}}Fuse\\ Function\end{tabular}} &
  \textbf{\begin{tabular}[c]{@{}c@{}}Training \\ Accuracy\\ {[}\%{]}\end{tabular}} &
  \textbf{\begin{tabular}[c]{@{}c@{}}Testing \\ Accuracy\\ {[}\%{]}\end{tabular}} & \textbf{\begin{tabular}[c]{@{}c@{}}Testing \\ F1 score\\ {[}\%{]}\end{tabular}} \\[1ex] \hline \hline
\textbf{12} & 96.56 & 89.82 & 71.85        \\[1ex]
\textbf{3}  & 97.45 & 89.84 & 71.88        \\[1ex]
\textbf{7}  & 97.47 & 89.86 & 71.91        \\[1ex]
\textbf{2}  & 96.58 & 89.96 & 71.93        \\[1ex]
\textbf{11} & 96.43 & \textbf{90.03} & \textbf{72.00}
\end{tabular}\label{tab:horizontal}
\end{table}

As can be seen in the results, while using the same network training recipe, our model outperformed the best results of ResNet50 while maintaining a comparable parameter count. In addition, as a slight overfitting of the model is apparent, by using a different learning rate scheduler or additional regularization, perhaps the result obtained could be even higher.

\headline[Vertical Fusion]

For vertical fusion experiments, we used the implementation of our \textit{Fuse\_V} architecture. Similarly to \textit{Fuse\_H} it had 4 separate feature extraction stages for \textit{Tensor A} and \textit{Tensor B} with feature fusion performed with features from two consecutive stages. Two more feature extraction layers were also added after the final concatenation of features. For the final classification, we used 4-layer MLP. Again, for nonlinearities, the ReLU function was used.

The network had a total of $29.52$ M parameters -- slightly more than \textit{Fuse\_H} model, but still, the number is comparable with ResNet50 and much lower than VGG or ViT networks. In Tab.~\ref{tab:vertical} we show 5 best performing models with their feature fusion method.

The results obtained show similar accuracies to \textit{Fuse\_H} model, however, \textit{Fuse\_V} models show better training patterns with overfitting less apparent. It is also important to notice, that this time again, we obtained better results than when using predefined models such as ResNet or ViT while maintaining the same training recipe. 

\begin{table}[]
\centering
	\caption{Five best performing methods in our \textit{Fuse\_V} model}
	\vspace*{2pt}
\begin{tabular}{P{1.2cm}|P{1.7cm}P{1.7cm}P{1.7cm}}
\textbf{\begin{tabular}[c]{@{}c@{}}Fuse\\ Function\end{tabular}} &
  \textbf{\begin{tabular}[c]{@{}c@{}}Training \\ Accuracy\\ {[}\%{]}\end{tabular}} &
  \textbf{\begin{tabular}[c]{@{}c@{}}Testing \\ Accuracy\\ {[}\%{]}\end{tabular}} & \textbf{\begin{tabular}[c]{@{}c@{}}Testing \\ F1 score\\ {[}\%{]}\end{tabular}} \\[1ex] \hline \hline
\textbf{9} & 95.95 & 89.79 &   71.83      \\[1ex]
\textbf{3}  & 98.28 & 89.80 &  71.85       \\[1ex]
\textbf{2}  & 96.39 & 89.82 &  71.85       \\[1ex]
\textbf{11}  & 95.11 & 89.97 & 71.96        \\[1ex]
\textbf{8} & 94.48 & \textbf{90.04} & \textbf{71.99}
\end{tabular}\label{tab:vertical}
\end{table}

\headline[Two-way fusion]

In the last part of the feature fusion functions evaluation, we combined both fusion modules and developed \textit{Fuse\_HV} model. For feature extraction performed separately on \textit{Tensor A} and \textit{Tensor B}, 4 feature extraction stages were used. The two-way fusion meant that the features were not only fused between features coming from the same network levels with \textit{Fuse Module H}, but in the next step, also linked backward after every two levels of the network with \textit{Fuse Module V}. Due to this, we did not have to perform any concatenation operation before the last two stages of feature extraction, keeping the dimensions of the feature tensor low. The classifier was again a 4-layer MLP. 

Thanks to the lack of concatenation, the network had a total of only $8.15$ M parameters which is considerably smaller than all of the networks from baseline experiments. In Fig.~\ref{fig:fuse_comparison} we show the full matrix of accuracies we obtained with cross-analysis performed, and in Tab.~\ref{tab:hor_ver} we also show 5 best-performing models with fuse functions listed.

\begin{table}[]
\centering
	\caption{Five best performing methods in our \textit{Fuse\_HV} model}
	\vspace*{2pt}
\begin{tabular}{P{1.6cm}|P{1.5cm}P{1.5cm}P{1.5cm}}
\textbf{\begin{tabular}[c]{@{}c@{}}Fuse\\ Function\end{tabular}} &
  \textbf{\begin{tabular}[c]{@{}c@{}}Training \\ Accuracy\\ {[}\%{]}\end{tabular}} &
  \textbf{\begin{tabular}[c]{@{}c@{}}Testing \\ Accuracy\\ {[}\%{]}\end{tabular}} & \textbf{\begin{tabular}[c]{@{}c@{}}Testing \\ F1 score\\ {[}\%{]}\end{tabular}} \\[1ex] \hline \hline
\textbf{H: 7, V: 11} & 94.83 & 89.92 & 71.92        \\[1ex]
\textbf{H: 11, V: 12}  & 93.62 & 89.93 & 71.96        \\[1ex]
\textbf{H: 12, V: 12}  & 93.05 & 89.96 & 71.96        \\[1ex]
\textbf{H: 6, V: 7}  & 93.64 & 89.97 & 71.98        \\[1ex]
\textbf{H: 11, V: 8} & 93.06 & \textbf{90.05} & \textbf{72.04}
\end{tabular}\label{tab:hor_ver}
\end{table}

\Figure[t!](topskip=0pt, botskip=0pt, midskip=0pt)[width=0.80\linewidth]{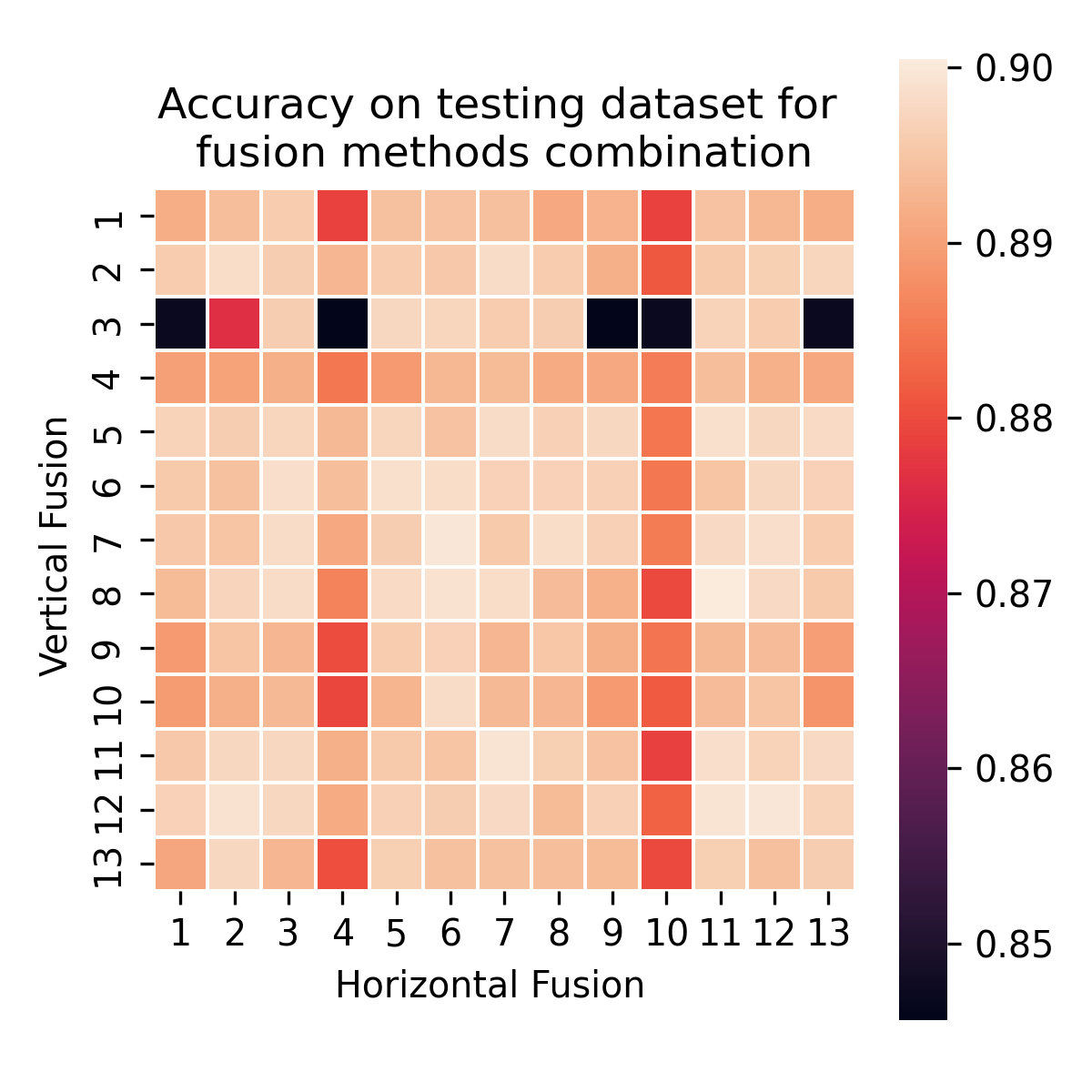}
{Cross-comparison of different fuse functions used for horizontal and vertical fusion in the same model on IDA-BD dataset.\label{fig:fuse_comparison}}

As can be seen in Tab.~\ref{tab:hor_ver}, despite having nearly 3 times fewer parameters than ResNet50, our model still managed to outperform it in both testing accuracy and F1 score. Employing two-way fusion also helped with overfitting, as \textit{Fuse\_HV} shows less overfitting than \textit{Fuse\_H} and \textit{Fuse\_V} models. Also, the table shows consistency with previous experiments for the best performing model, as both vertical and horizontal fusion functions are the methods best performing in \textit{Fuse\_H} and \textit{Fuse\_V} respectively, indicating that those are the best fitting methods for the operation performed.

Furthermore, in Fig.~\ref{fig:fuse_comparison} we can see, that the worst performing vertical fusion functions are \textbf{3} and \textbf{4}, while the worst horizontal fusion functions were \textbf{4} and \textbf{10}. For this reason, in our further experiments, those methods were ultimately excluded.

To delve more deeply into the results we obtained, in Fig.~\ref{fig:missclassifications} we show some of the image pairs misclassified by the network. For \textbf{IDA-BD} dataset, example A) shows a common practice of securing damaged roofs with blue tarp after the hurricane -- this blue coating can severly mask the actual damage to the building, so that the true condition of the building is only signaled by the presence of the coverage and its size. In example B) another problem is shown -- some of the buildings were on the edge of the images, making some part of it of no significance to the network. In \textbf{xView2} dataset, example C) shows the problem of object occlusion, typically with extensive foliage (as in the example) or clouds which is similar to example A). In example D) the problem of small buildings and slight shift in perspective is apparent. This problem however can be tackled by adding more sorrounding area to the image for classification. 

\Figure[t!](topskip=0pt, botskip=0pt, midskip=0pt)[width=0.65\textwidth]{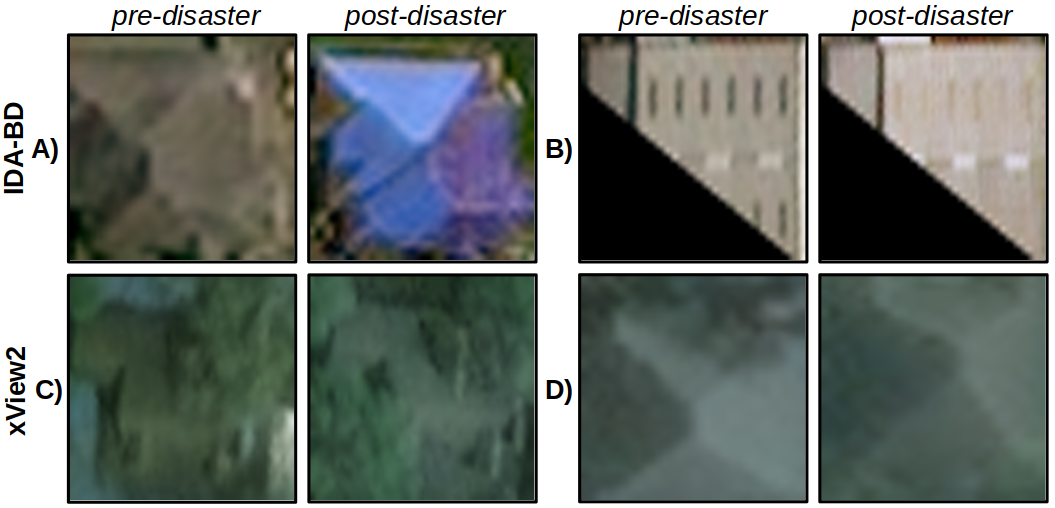}
{Some of examples of missclassified pairs of images from both dattasets used.\label{fig:missclassifications}}

\subsection{Fuse Module in predefined models}\label{sec:MSFF_eval}

For our last experiments, we used our \textit{Fuse Module} in predefined network architectures in order to show the versatility of our method. For the experiments, we used our \textit{Fuse\_HV} model as the reference as well as modified ResNet50 and ViT B architectures and tested them on both \textbf{IDA-BD} and \textbf{xView2} datasets. We also maintained our previous training recipe, with ADAM optimizer, \textit{Reduce on Plateau} learning rate scheduler, and 60 training epochs for the fairness of comparison.

In order to provide a more fair comparison with baseline experiments, we decided to add only one \textit{Fuse Module} in each architecture in the place of concatenation operation. Furthermore, to verify the relevance of two separate feature extraction paths from \textit{Tensor A} and \textit{Tensor B}, we prepared models in two variants. In the first variant, both feature tensors are extracted with the same layers of feature extractors and then fused with the \textit{Fuse Module} before classification MLP -- models \textit{ResNet50$_{S}$} and \textit{ViT$_{S}$} (\textit{S} stands for \textit{single} in the term of feature extraction paths). In the second variant, features from both tensors are extracted separately, with their own copy of individually trained feature extractors, and then fused with the \textit{Fuse Module} -- models \textit{ResNet50$_{D}$} and \textit{ViT$_{D}$} (\textit{D} for \textit{double} feature extraction paths). These models will differ in the number of trainable feature extractor parameters, but the final layers belonging to the MLP classifier will remain the same.

In Tab.~\ref{tab:final} we show the best-performing models in our experiments with their parameters count for both \textbf{IDA-BD} and \textbf{xView2} datasets along with the feature fusion method that was used.

\begin{table}[]
\centering
\caption{Predefined models experiments results}
\vspace*{2pt}
\begin{tabular}{c|cccc}
\textbf{\begin{tabular}[c]{@{}c@{}}Network\\ / \\ Fuse Function\end{tabular}} &
  \textbf{Dataset} &
  \textbf{\begin{tabular}[c]{@{}c@{}}Training\\ Accuracy\\ {[}\%{]}\end{tabular}} &
  \textbf{\begin{tabular}[c]{@{}c@{}}Testing\\ Accuracy\\ {[}\%{]} / \\ F1 score {[}\%{]}\end{tabular}} &
  \textbf{\begin{tabular}[c]{@{}c@{}}Parameter\\ Count {[}M{]}\end{tabular}} \\[1ex] \hline \hline 
\multirow{2}{*}{\textbf{\begin{tabular}[c]{@{}c@{}}Fuse\_HV\\ H: 11, V: 8\end{tabular}}}    & IDA-BD & 93.06 & \textbf{90.05} / \textbf{72.01}         & \multirow{2}{*}{\textbf{8.15}}  \rule[-2ex]{0pt}{5ex}       \\
                                                                              & xView2 & 90.73 & 89.64 / 71.91&                                 \\ \hline
\multirow{2}{*}{\textbf{\begin{tabular}[c]{@{}c@{}}ResNet50$_{S}$\\ H: 8\end{tabular}}} & IDA-BD & 94.68 & 89.45 / 71.88        & \multirow{2}{*}{23.52} \rule[-2ex]{0pt}{5ex} \\
                                                                              & xView2 & 90.82 & 88.97 / 71.82        &                                 \\ \hline
\multirow{2}{*}{\textbf{\begin{tabular}[c]{@{}c@{}}ResNet50$_{D}$\\ H: 11\end{tabular}}} & IDA-BD & 94.56 & 89.92 / 71.96 & \multirow{2}{*}{47.02}    \rule[-2ex]{0pt}{5ex}      \\
                                                                              & xView2 & 91.30 & \textbf{89.77} / \textbf{71.99}         &                                 \\ \hline
\multirow{2}{*}{\textbf{\begin{tabular}[c]{@{}c@{}}ViT$_{S}$\\ H: 11\end{tabular}}}     & IDA-BD & 97.97 & 88.65 / 71.01        & \multirow{2}{*}{85.80}     \rule[-2ex]{0pt}{5ex}     \\
                                                                              & xView2 & 91.34 & 89.39 / 71.69         &                                 \\ \hline
\multirow{2}{*}{\textbf{\begin{tabular}[c]{@{}c@{}}ViT$_{D}$\\ H: 11\end{tabular}}}     & IDA-BD & 95.36 & 88.90 / 71.23        & \multirow{2}{*}{171.60}    \rule[-2ex]{0pt}{5ex}     \\
                                                                              & xView2 & 91.48 & 89.46 / 71.74        &                                
\end{tabular}\label{tab:final}
\end{table}

As can be seen in the results, using the \textit{Fuse Module} helped to obtain considerably higher accuracies and F1 scores in all of the models. The effect is most apparent for ViT B architecture where using our method, the accuracy increased by more than 3 percentage points. Furthermore, all the best-performing networks again used the best fusion methods (method \textbf{8} or \textbf{11}) identified in Section~\ref{sec:evaluation} and not only increased the models' performance but also reduced the overfitting during the training. However, still considering the total parameter count of the models, the best performing one is \textit{Fuse\_HV}, having only slightly lower testing performance on \textbf{xView2} dataset than ResNet50$_{D}$ model. This thus provides a rationale for using a combination of horizontal and vertical modules in our further work on this method with predefined models.

To summarize our \textit{Fuse Module} experiments, it can be seen that the key feature of the \textit{Fuse Module} is the ability to maintain the initial dimensions of the input tensor so that as opposed to performing fusion with a concatenation operation, the network itself keeps a low parameter count without the need for supplementary downsampling layers. Additionally, it also allows for using it with pre-trained networks without interfering with their architecture and increasing their parameter count. It is also apparent, that using the \textit{Fuse Module} can increase the network performance, regardless of its initial architecture, however further work should be considered to combat class imbalance in the datasets.

\section{Conclusions and future work}\label{sec:conclusions}

This paper presents a novel method of feature fusion that can be employed in both vertical and horizontal data flow in the CNN network in the task of data comparison and verification. The proposed method shown to improve the effectiveness of small CNN models as well as state-of-the-art architectures, including ViT transformer models.

In the paper, we proposed multiple feature fusion techniques and verified them on two datasets -- \textbf{IDA-BD} and \textbf{xView2} containing satellite images for structure damage state assessment after natural disasters. In our experiments, we combined vertical and horizontal fusion on multiple models, including our own implementation of a low-parameter count CNN network, that ultimately achieved better performance than much larger models. 

With our further work, we intend to expand our experiments with predefined models, by upgrading them with multiple \textit{Horizontal Fusion Modules} after each stage of feature extraction, but also by adding additional \textit{Vertical Fusion Modules} that proved to be effective in our experiments. Also, worth considering is the usage of initially pretrained models (\textit{e.g.} with ImageNet dataset) as it tends to increase the performance of the models.

We intend to explore other tasks and datasets that could benefit from our method as well. One of them is security verification in the task of entity verification for \textit{e.g.} human face or fingerprint recognition. In such applications, we could employ our two-branch network for one-to-many verification check and explore its ability to distinguish familiar entities in a setup of assessing the certainty of known/unknown person in the data from outside the initial dataset of known entities. Moreover, by using recurrent neural networks and language-oriented transformer models, another topic worthy of exploring could be the task of sequence similarity search for finding similar proteins in archive datasets -- a task that is currently performed with Basic Local Alignment Search Tool or Many-against-Many sequence searching. Lastly, by employing other datasets, we also intend to explore domain shift scenarios and cross-dataset transfer learning.

\def\bibfont{\footnotesize}
\bibliographystyle{IEEEtranN}
\bibliography{bib}{}

\begin{IEEEbiography}[{\includegraphics[width=1in,height=1.25in,clip,keepaspectratio]{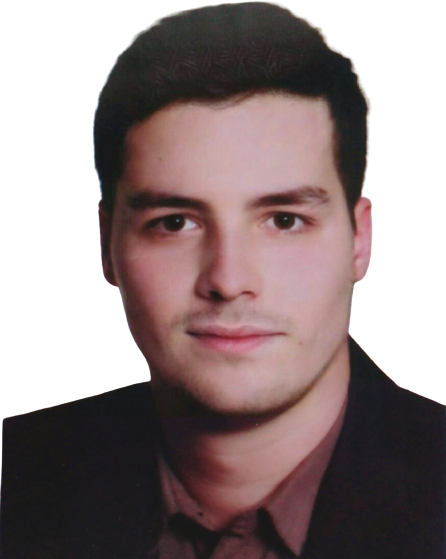}}]{Mateusz \.Zarski} received his PhD degree from the Silesian University of Technology in 2023 with his dissertation on applied Computer Vision and Deep learning in civil engineering-related issues.

From 2021, he works at the Institute of Theoretical and Applied Informatics, Polish Academy of Sciences, Gliwice, Poland, and after obtaining PhD, works there as an assistant professor. His research interest includes applied Artificial Intelligence in the fields of civil engineering, medical imaging and Reinforced Learning.
\end{IEEEbiography}

\begin{IEEEbiography}[{\includegraphics[width=1in,height=1.25in,clip,keepaspectratio]{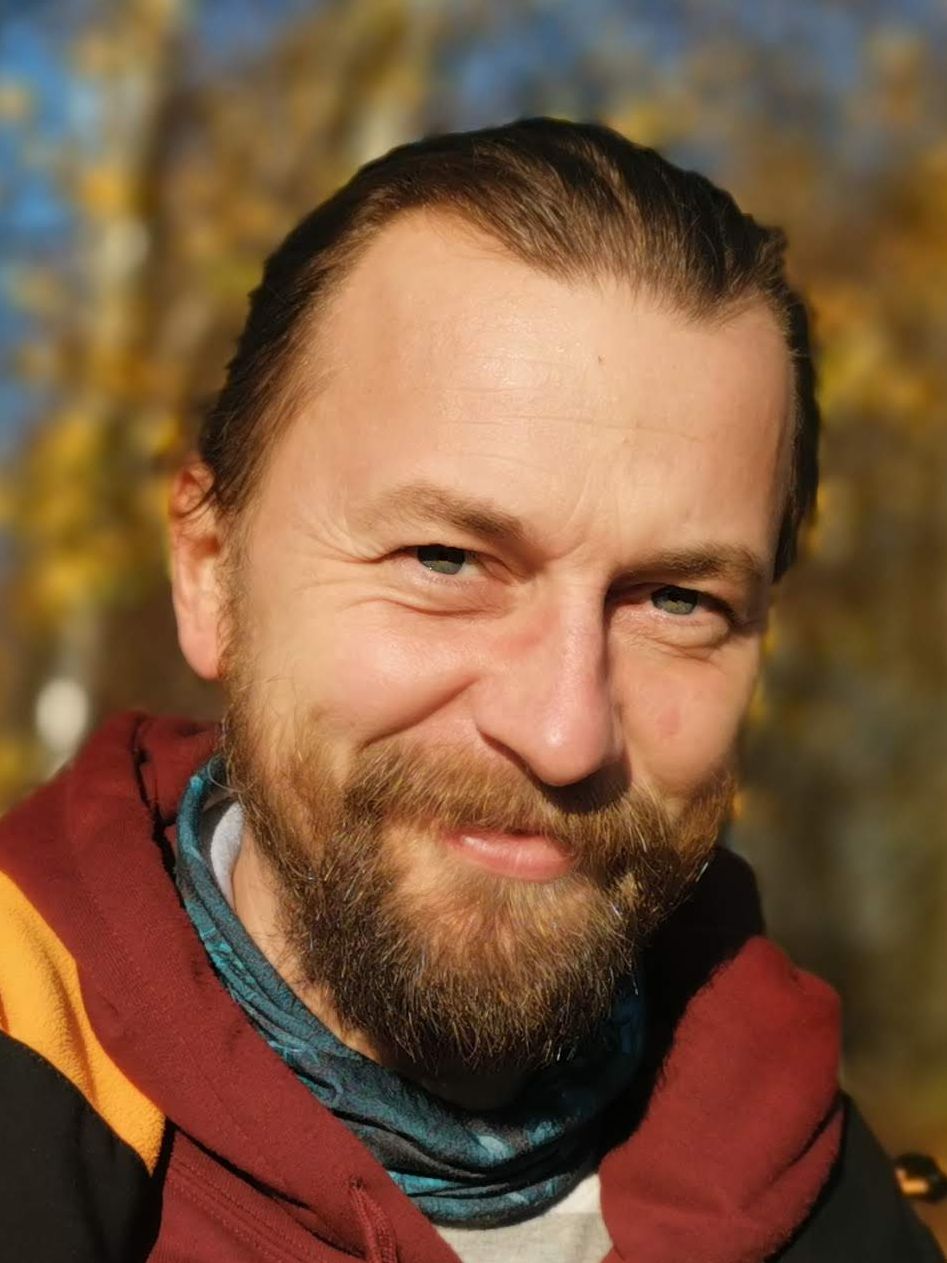}}]{Jaroslaw Miszczak} received his PhD degree from the Institute of Theoretical and Applied Informatics, Polish Academy of Sciences, Gliwice, Poland in 2008, and DSc from the Silesian University of Technology in 2014.

He works as an associate professor at the Institute of Theoretical and Applied Informatics, Polish Academy of Sciences.  His research interests include quantum computing, scientific computing, and complex systems.
\end{IEEEbiography}

\EOD

\end{document}